\documentclass[conference]{IEEEtran}
\IEEEoverridecommandlockouts
\usepackage{cite}
\usepackage{amsmath,amssymb,amsfonts}
\usepackage[ruled]{algorithm2e}
\usepackage{graphicx}
\usepackage{textcomp}
\usepackage{authblk} 
\usepackage{xcolor}
\usepackage{subcaption}
\usepackage{float}

\def\BibTeX{{\rm B\kern-.05em{\sc i\kern-.025em b}\kern-.08em
    T\kern-.1667em\lower.7ex\hbox{E}\kern-.125emX}}
\begin{document}

\title{RE-POSE: Synergizing Reinforcement Learning-Based Partitioning and Offloading for Edge Object Detection}

\author[12]{Jianrui Shi}
\author[2]{Yong Zhao\textsuperscript{\dag}\thanks{\dag Corresponding author: Yong Zhao. Email: y1zhao@polyu.edu.hk}}
\author[2]{Zeyang Cui}
\author[2]{Xiaoming Shen}
\author[23]{Minhang Zeng}
\author[4]{Xiaojie Liu}

\affil[1]{Department of Electrical and Computer Engineering, National University of Singapore, Singapore}
\affil[2]{Department of Computing, The Hong Kong Polytechnic University, Hong Kong}
\affil[3]{School of Computing and Data Science, The University of Hong Kong, Hong Kong}
\affil[4]{Pengcheng Laboratory, Shenzhen, China}


\maketitle

\begin{abstract}
    Object detection plays a crucial role in smart video analysis, with applications ranging from autonomous driving and security to smart cities. However, achieving real-time object detection on edge devices presents significant challenges due to their limited computational resources and the high demands of deep neural network (DNN)-based detection models, particularly when processing high-resolution video. Conventional strategies, such as input down-sampling and network up-scaling, often compromise detection accuracy for faster performance or lead to higher inference latency. To address these issues, this paper introduces RE-POSE, a Reinforcement Learning (RL)-Driven Partitioning and Edge Offloading framework designed to optimize the accuracy-latency trade-off in resource-constrained edge environments. Our approach features an RL-Based Dynamic Clustering Algorithm (RL-DCA) that partitions video frames into non-uniform blocks based on object distribution and the computational characteristics of DNNs. Furthermore, a parallel edge offloading scheme is implemented to distribute these blocks across multiple edge servers for concurrent processing. Experimental evaluations show that RE-POSE significantly enhances detection accuracy and reduces inference latency, surpassing existing methods.

\end{abstract}


\section{Introduction}
With rapid advancements in Deep Neural Networks (DNNs) and the widespread deployment of camera-equipped devices, smart video analysis has become essential in applications like autonomous driving, security surveillance, smart cities, and robot navigation \cite{gupta2021deep}. Central to this technology is object detection, which enables real-time detection and tracking of objects, supporting functionalities such as situational awareness and automated responses \cite{yang2022flexpatch}.

Deploying smart video analysis at the edge offers significant advantages over cloud-based solutions, including reduced latency and enhanced privacy by processing data locally \cite{mao2017survey}. However, edge devices often have limited computational and storage resources, making it hard to run resource-intensive DNN-based object detection models on high-resolution video streams. High-resolution inputs require substantial computational power for real-time inference, and current mitigation strategies—like down-sampling or using smaller models—typically trade off accuracy for reduced latency \cite{zhao2017random, jiang2021flexible}.

To address these challenges, various model compression techniques have been developed, including weight and branch pruning \cite{guo2020multi}, weight sharing \cite{gao2019cross}, tensor quantization \cite{gupta2022compression}, knowledge distillation \cite{gou2021knowledge}, and network architecture search \cite{sukthanker2024large}. While these methods effectively reduce model size for edge deployment, they often results in decreased accuracy.

Additionally, several acceleration methods targeting video tasks have been proposed. Some methods, like \cite{shi2023adapyramid, zhang2021elf}, divide high-resolution images into smaller blocks, processing only those likely to contain pedestrians. However, \cite{shi2023adapyramid} has limitations such as outdated performance estimations, inadequate background removal in crowded areas, and overly fine partitions that degrade accuracy. Similarly, Elf \cite{zhang2021elf} uses an attention-based LSTM to predict pedestrian locations, but it can be slow in densely populated scenes due to multiple inferences.

To overcome these limitations, we introduce \textbf{RE-POSE} (Reinforcement Learning-Driven Partition-Offloading Synergy at Edge), a framework that optimizes the balance between detection accuracy and inference latency in resource-constrained edge environments. RE-POSE leverages Reinforcement Learning (RL) to adaptively partition high-resolution video frames into blocks and offload them to multiple edge servers for parallel inference, improving detection performance while maintaining latency constraints.

Our main contributions are as follows:

\begin{itemize}
    \item \textbf{RE-POSE Framework}: Presents a novel RL-driven architecture that performs non-uniform partitioning and parallel offloading, achieving superior real-time object detection performance at edge.
    
    \item \textbf{RL-Based Dynamic Clustering Algorithm (RL-DCA)}: Dynamically partitions video frames into non-uniform blocks based on coarse object detection and RL, enhancing detection accuracy and computational efficiency.
    
    \item \textbf{Parallel Edge Offloading Scheme}: Efficiently assigns image blocks to multiple edge servers, selectively processing blocks with objects to minimize inference latency.
\end{itemize}

The remainder of this paper is organized as follows. Section \ref{sec:system overview} describes the RE-POSE framework. In Section \ref{sec:Proposed Methodology}, we detail the RE-POSE framework, including the RL-DCA and parallel edge offloading schemes. Section \ref{sec:performance evaluation} provides a comprehensive performance evaluation, and Section \ref{sec:conclusion} concludes the paper.

\section{System Overview}
\label{sec:system overview}

\begin{figure*}[th]
    \centering
    \begin{minipage}{0.85\textwidth}
        \includegraphics[width=\textwidth]{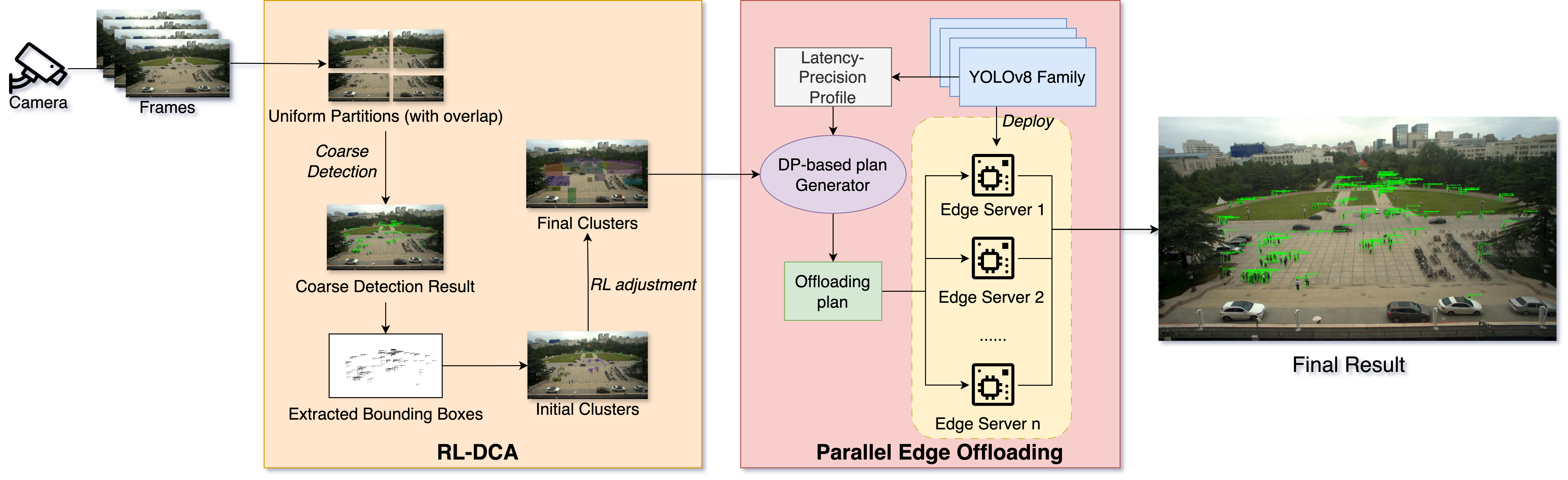}
    \caption{RE-POSE framework overview}
    \label{fig:REPOSE framework}
    \end{minipage}
\end{figure*}

The RE-POSE framework adopts RL-driven partitioning and parallel offloading to achieve accurate object detection on edge devices. It includes two main modules: \textbf{RL-DCA Partitioning} and \textbf{Parallel Edge Offloading}, shown in Fig. \ref{fig:REPOSE framework}.

\subsection{RL-DCA Partitioning}
\label{subsubsec:RL-DCA Partitioning}
The input is a high-resolution video frame, initially divided into uniform blocks for coarse detection using YOLOv8 model with a low confidence threshold to ensure high recall. Detected objects are clustered with MeanShift to form initial clusters. RL-DCA then refines these clusters using a PPO-based RL agent that dynamically merges or splits clusters to achieve compact, homogeneous groupings. This adaptive clustering step ensures that each cluster can be handled by a suitable model without unnecessary computational overhead. (\ref{subsec:Partitioning Scheme})

\subsection{Parallel Edge Offloading}
\label{subsubsec:Parallel Edge Offlaoding}
After clustering, a bounding box (i.e, image block) is generated for each cluster, including all detection boxes within it. Using these bounding boxes, a Dynamic Programming (DP)-based plan generator uses a latency-precision profile of various YOLOv8 models to assign the optimal model to each block. This selection maximizes accuracy under a global latency constraint. The selected models are offloaded in parallel to multiple edge servers, enabling parallel processing. Finally, the detected objects from each cluster are aggregated into the final output. (\ref{subsec:task offloading scheme})

\section{Proposed Methodology}
\label{sec:Proposed Methodology}

\subsection{Partitioning Scheme}
\label{subsec:Partitioning Scheme}
This section presents our RL-based Dynamic Clustering Algorithm (RL-DCA). Unlike existing methods \cite{zhang2021elf, shi2023adapyramid, jiang2021flexible} that rely on predefined parameters or heuristics, RL-DCA is fully self-adaptive. RL-DCA can perform reasonable partition of the current frame without relying on any historical frame information. This is because the RL agent learns rich representations to handle diverse object scenarios in the training phase. In the following subsections, we describe the components and functioning of RL-DCA.

\subsubsection{Coarse Detection}
\label{subsubsec: Coarse Detection}
For each input video frame, we first partition the frame into $n \times E$ equally sized blocks, where $E$ represents the number of edge devices in the system. This partitioning strategy ensures that each edge device processes a smaller portion of the frame, thereby evenly distributing the computational load and fully leveraging the collective computational power of the edge devices. Furthermore, after partitioning, the size of the objects in each block increases relative to the block resolution, making them more prominent in the feature maps. This improvement in object-to-pixel ratio enhances the detection accuracy and minimizes the risk of missing objects. Next, coarse detection is performed on the partitioned blocks. To maximize recall and ensure no objects are overlooked, we lower the confidence threshold appropriately. Missed detections, which are often caused by severe occlusion in densely packed areas, are mitigated in subsequent stages through clustering processes that leverage surrounding detections to recover the missed objects. Detection results from the edge devices are aggregated using Non-Maximum Suppression (NMS) to produce a unified set of object detections.

\subsubsection{Initial Clustering}
\label{subsubsec: Initial Clustering}
Following the aggregation of detection results, RL-DCA uses the MeanShift algorithm for initial clustering of detected objects. MeanShift is a density-based method that identifies clusters of arbitrary shapes without requiring a predefined number of clusters, making it ideal for varying object distributions. This initial clustering organizes objects based on their spatial locations, forming preliminary clusters that serve as a foundation for further refinement by the RL agent.

To address the stratified distribution of object sizes in high-resolution frames—where upper blocks contain smaller, densely packed objects and lower blocks have larger, sparsely distributed objects—we apply a non-linear transformation to the y-coordinates of detected objects:
\begin{equation}
    y_T = y^{\alpha_T}
\end{equation}
where $y$ is the normalized y-coordinate of the object center in the range $[0, 1]$, $\alpha_T$ is a tunable parameter $(0<\alpha_T<1)$ that controls the degree of transformation for the y-values. Stretching y-coordinates in the upper blocks improves separability, while compressing them in the lower blocks prevents unnecessary splitting of larger objects.

\subsubsection{PPO-based Cluster Adjustment}
\label{subsubsec: Cluster Adjustment}

To address the limitation of MeanShift clustering, which solely considers the center points of detection boxes, we integrate RL policy network into our clustering pipeline.

Specifically, we employ the Proximal Policy Optimization (PPO) \cite{schulman2017proximal} algorithm, an actor-critic policy gradient method renowned for its balance between performance and computational efficiency. In this framework, the actor (policy network) interacts with the environment by selecting actions, while the critic network estimates the value function, predicting the expected cumulative rewards from given states.

\paragraph{Algorithm Overview}
\label{pg: Algorithm Overview}
In RL-DCA, the policy network selects actions (keep, merge, or split clusters) based on the current clustering state, while the critic network evaluates the resulting state-action sequence to provide value estimates. By incorporating a custom reward function that quantifies clustering quality, PPO guides the policy network to learn optimal clustering adjustment strategies. Below, we detail the integration of PPO into RL-DCA, including the mathematical formulations of state (\ref{pg: State Representation}), action (\ref{pg: Action Space}), and reward (\ref{pg: Reward Function}).

\paragraph{State Representation}
\label{pg: State Representation}
The state representation encodes the current clustering configuration $\mathcal{C}=\left \{ C_1, C_2,\cdots, C_N\right \}$ as a feature vector $s$. This state vector provides the actor and critic networks with sufficient information to make decisions and evaluate them. It has following features:
\begin{itemize}
    \item \textbf{Centroids} $(\mu_x, \mu_y)$: Spatial center of each cluster.
    \item \textbf{Average Object Size} $(\mu_w, \mu_h)$: Mean width and height of the bounding box in the cluster.
    \item \textbf{Cluster Size} $(S_i)$: Number of objects within each cluster.
    \item \textbf{Cluster Count} $(N)$: Total number of clusters.

\end{itemize}
The state vector for $N$ clusters at step $t$ is represented as:
\begin{equation}
\begin{split}
     s_t = [\mu_x^{(1)}, \mu_y^{(1)}, \mu_w^{(1)}, \mu_h^{(1)}, S_1, \cdots, \\
      \mu_x^{(N)}, \mu_y^{(N)}, \mu_w^{(N)}, \mu_h^{(N)}, S_N]
\end{split}
\end{equation}
As the input of policy network, the state will be padded or truncated to a fixed size to ensure consistency.

\paragraph{Action Space}
\label{pg: Action Space}
The policy network in RL-DCA selects one of the following discrete actions $a_t$ based on the current state $s_t$:

    \textbf{Keep}: Maintain the current clustering configuration without changes. We define it as $a_{keep}$. As a result:
    \begin{equation}
        s_{t+1} = s_t,\quad\mathcal{C}_{t+1} = \mathcal{C}_t
    \end{equation}
    This action does not alter the state vector and is used when the clustering configuration is already optimal.
    
    \textbf{Merge}: The actor chooses two clusters closest in centroid distance to merge, this approach reduces redundant clusters and ensures that closely positioned objects are grouped together.
    The pair with the minimum distance is selected for merging, we define the merge action as: 
    \begin{equation}
        a_{merge}(C_i, C_j) \quad where \ (i, j) = \mathop{\arg\min}\limits_{i\ne j}\ d_{ij}
    \end{equation}
    The state transition and clustering configuration change for merge action are:
    \begin{equation}
        s_{t+1} = {\rm Update}(s_t, a_{merge}(C_i, C_j))
    \end{equation}
    \begin{equation}
        \mathcal{C}_{t+1} = \left \{\mathcal{C}_t \setminus\left \{C_i, C_j\right \}\right \}\cup\left \{C_{new}\right \}
    \end{equation}
    
    \textbf{Split}: Select a cluster $C_i$ and divide it into two sub-clusters $C_{new1}$ and $C_{new2}$. The process involves analyzing the spatial distribution of object centers within the cluster and applying 1D K-Means clustering along the most varied dimension (x-axis or y-axis). We define the split action as $a_{split}(C_i)$.

    \textit{Why 1D K-Means?} (i)As observed in \ref{subsubsec: Initial Clustering}, objects in the vertical (y-axis) dimension often exhibit significant variations in size and density, using 1D K-Means clustering along the y-axis can effectively separate clusters with vertically distinct object sizes; (ii)Object detection algorithms process rectangular blocks as inputs. By splitting clusters along a single dimension, the overlap between resulting sub-clusters is minimized; (iii)1D K-Means is computationally simpler than 2D K-Means.

    We first compute the variance of detection box centers in both x- and y-directions,
    then select the dimension with the larger variance as the splitting direction $D$:
    \begin{equation}
        D = \mathop{\arg\min}\limits_{d \in {\left \{ x, y \right \}}}\ {\rm Var_d}
    \end{equation}
    the clustering process is:
    {\small
    \begin{equation}{
        \left \{ C_{new1}, C_{new2} \right \} = {\rm K\text{-}Means}(C_i, k = 2, direction=D)}
    \end{equation}
    }
        
    where $k=2$ specifies that the cluster will be split into two sub-clusters. The new cluster information including bounding box distribution $x_k,y_k,w_k,h_k$ and size $S_{new1}$ and $S_{new2}$ can be obtained after the K-Means process.
    
    The state transition and clustering configuration change for split action are:
    \begin{equation}
        s_{t+1} = {\rm Update}(s_t, a_{split}(C_i))
    \end{equation}
    \begin{equation}
        \mathcal{C}_{t+1} = \left \{\mathcal{C}_t \setminus\left \{C_i\right \}\right \}\cup\left \{C_{new1}, C_{new2}\right \}
    \end{equation}

After all the discussion, the action space $A$ can be expressed as:
\begin{equation}
    A = \left \{a_{keep}\right \} \cup \left \{a_{merge}(C_i, C_j)\right \} \cup \left \{a_{split}(C_i)\right \}
\end{equation}

\paragraph{Reward Function}
\label{pg: Reward Function}
At each decision step, after the actor selects an action and the environment updates the clustering configuration accordingly, a reward is computed, which guides the PPO agent to improve cluster quality over time.

This reward is based on multiple metrics that reflect the desired properties of the clusters:

    \textbf{Cluster Tightness Penalty}: For each cluster $C_i$, compute the mean distance of all its objects to the cluster centroid $(\mu_x,\mu_y)$, then average this quantity over all $N$ clusters:
    \begin{equation}
        R^{(1)} = -\frac{1}{N}\sum_{i=1}^N\frac{1}{S_i}\sum_{(x,y)\in C_i}\left \| (x, y) - (\mu_x^{(i)}, \mu_y^{(i)}) \right \|
    \end{equation}
    This ensures that the objects within a cluster are compactly grouped in the sub-image, reducing the proportion of background and enhancing detection accuracy. 
    
    By penalizing intra-cluster distances, the agent is incentivized to perform splits that yield more compact clusters, as much splits always reduce this penalty.
    
    \textbf{Object Area Variance Penalty}: For each cluster $C_i$, compute the variance of object areas $A_{ik}$, then average these variances across all clusters:
    \begin{equation}
        R^{(2)} = -\frac{1}{N}\sum_{i=1}^N{\rm Var}(A_{i1},A_{i2},\cdots,A_{iS_i})
    \end{equation}
    This ensures that objects within the same cluster have relatively uniform bounding box areas, minimizing heterogeneity in object sizes. Since the size of bounding boxes is often strongly correlated with the y-axis position, this penalty inherently encourages vertical splits for clusters with significant size variation.
    
    By penalizing object area variance, the agent is encouraged to group objects of similar size together, possibly by splitting large, diverse clusters into more homogeneous sub-clusters or merging smaller, uniformly sized object clusters.
    
    \textbf{Cluster Count Penalty}: A penalty is applied if the number of clusters $N$ falls outside the acceptable range $(N_{min}, N_{max})$, increasing proportionally to the deviation from the minimum or maximum limits. No penalty is applied if $N$ is within the range:
    \begin{equation}
        R^{(3)}=\left\{\begin{array}{ll}
        -\left(N_{min}-N\right) & \text { if } N<N_{min} \\
        -\left(N-N_{max}\right) & \text { if } N>N_{max} \\
        0 & \text { otherwise }
    \end{array}\right.
    \end{equation}

    This reward means to keep the number of clusters $N$ within a desirable range. This ensures that the clustering is neither too coarse (too few clusters) nor too fragmented (too many clusters), helps stabilize the clustering configuration, preventing extreme actions that would produce too many or too few clusters. The agent learns to split or merge just enough times to keep $N$ within $(N_{min}, N_{max})$.
    
    \textbf{Close Cluster Penalty}: For each pair of distinct clusters $(C_i, C_j)$, check if the distance between their centroids is less than $d_m$, each pair that violates the separation threshold $\theta$ adds to the penalty: 
    {\small
    \begin{equation}
        R^{(4)}=-\frac{1}{2} \sum_{i=1}^{N} \sum_{\substack{j=1 \\ j \neq i}}^{N} \mathbb{I}\left(\left\|(\mu_{x}^{(i)}, \mu_y^{(i)})-(\mu_{x}^{(j)}, \mu_y^{(j)})\right\|<d_m\right)
    \end{equation}
    }
    
    The agent is encouraged to maintain adequate spacing between clusters, but more importantly, this penalty is designed to prevent the agent from repeatedly splitting the same cluster or its sub-clusters, leading to highly imbalanced cluster sizes and premature convergence once the desired number of clusters is reached. By discouraging excessively close clusters, the agent is incentivized to identify and prioritize clusters that genuinely require splitting, ensuring a more balanced and meaningful clustering configuration.

    \textbf{Combined Reward Function}: The final reward at time $t$ after taking an action is: \begin{equation}
        R_t = \alpha R^{(1)} + \beta R^{(2)} + \gamma R^{(3)} + \delta R^{(4)}
    \end{equation}
     The parameters $\alpha, \beta, \gamma, \delta$ serve as weighting factors for the respective components of the reward function. This comprehensive reward function guides the RL agent towards producing more meaningful, efficient, and accurate clustering configurations in RL-DCA.
\paragraph{Inference Stage}
Overall, RL-DCA dynamically refines clusters by merging or splitting them based on learned policies, as detailed in Algorithm~\ref{alg:rl-dca inference}. After obtaining the final clusters, the blocks including all the detection boxes within each cluster will be used as blocks for offloading tasks.
\begin{algorithm}[ht]
\caption{RL-DCA Inference}
\label{alg:rl-dca inference}
\LinesNumbered
\KwIn{
    $F$: input frame, 
    $n$: partition parameter, 
    $E$: number of edge devices, 
    $\alpha_T$: transform parameter, 
    $\pi_{\theta}$: policy network, 
    $T_{\max}$: max steps
}
\KwOut{
    $\mathcal{C}_{\text{final}}$: final clustering configuration
}

\textbf{Step 1: Partitioning and Detection}\\
Divide $F$ into $n \times E$ segments and perform object detection on each segment. Aggregate detections and apply NMS to obtain results $\mathcal{B}$.

\textbf{Step 2: Initial Clustering}\\
Transform y-coordinates using the formula:
$
y_T = y^{\alpha_T}
$
Apply MeanShift clustering to form initial clusters $\mathcal{C}^{(0)}$.

\textbf{Step 3: Cluster Adjustment}\\
\For{$t \gets 1$ \KwTo $T_{\max}$}{
    Compute state $s_t$ from current clusters $\mathcal{C}^{(t-1)}$\;
    Select action $a_t = \arg\max_{a}\pi_{\theta}(a|s_t)$\;
    Update clusters $\mathcal{C}^{(t)}$ based on $a_t$\;
}

\Return $\mathcal{C}_{\text{final}} = \mathcal{C}^{(T_{\max})}$\;
\end{algorithm}

\subsection{Task Offloading Scheme}
\label{subsec:task offloading scheme}
\subsubsection{Problem Formulation}

We formulate the model selection task within each partition in \ref{subsubsec:Parallel Edge Offlaoding} as a Multi-Choice Knapsack Problem (MCKP), aiming to maximize the overall detection accuracy under a global computational latency constraint. By modeling this decision-making process as an MCKP, we can systematically choose the most suitable object detection model for each partition given limited computational latency and diverse object characteristics.
\paragraph{Rationale for Partition-Level Model Selection}

The input image is partitioned into several blocks based on object size, distribution, and presence of the object. Using YOLOv8 models for object detection, these blocks are processed under specific computational constraints. Each YOLOv8 model \( j \) has a fixed processing latency \( d_j \) and accuracy \( p_j \) for different detection scales. For each block, the goal is to select a model that maximizes overall accuracy while adhering to a global computational latency constraint \( D_{\text{cap}}\).

\paragraph{Definitions}
For each bounding box cluster \( i \) (partition \( i \)) in the image:  
\begin{itemize}
    \item \( W_i, H_i \): Width and height of the original resolution of partition \( i \).  
    \item \( M_i \): Number of bounding boxes within cluster \( i \).  
    \item \( A_{ik} \): Original area of the \( k^{\text{th}} \) bounding box in partition \( i \).  
\end{itemize}

We consider five YOLOv8 models for selection: YOLOv8n $(640 \times 640)$, YOLOv8s $(768\times768)$, YOLOv8m $(896\times896)$, YOLOv8l $(1024\times1024)$ and YOLOv8x $(1280\times1280)$.
When model \( j \) processes partition \( i \), the image block is resized to \( S_j \times S_j \), and the bounding box area scales proportionally as:  
\begin{equation}
    A'_{ik}(j) = A_{ik} \times \frac{S_j^2}{W_i \times H_i}
\end{equation}
The precision mapping function \( p_j \) determines the accuracy corresponding to the resized area \( A'_{ik}(j) \). The average precision \( P_{ij} \) for partition \( i \) using model \( j \) is:  
\begin{equation}
    P_{ij} = \frac{1}{M_i} \sum_{k=1}^{M_i} p_j(A'_{ik}(j))
\end{equation}
\paragraph{Decision Variables}  
We define the binary decision variable \( x_{ij} \):  
\begin{equation}
    x_{ij} = 
\begin{cases} 
1 & \text{if model } j \text{ is selected for partition } i \\
0 & \text{otherwise}
\end{cases}
\end{equation}
\paragraph{Objective Function}
The objective is to maximize the overall accuracy across all partitions:  
\begin{equation}
    \max \sum_{i=1}^N \sum_{j} P_{ij} x_{ij}
\end{equation}
\paragraph{Constraints}  
\begin{itemize}
    \item \textbf{Model Selection for Each partition}: Each partition \( i \) must select exactly one model:  
    \begin{equation}
    \sum_{j} x_{ij} = 1, \quad \forall i = 1, \dots, N
    \end{equation}

    \item \textbf{Global Latency Constraint}: The total processing latency must not exceed the specified limit \( D_{\text{max}} \):  
    \begin{equation}
        \sum_{i=1}^N \sum_{j} d_j x_{ij} \leq D_{\text{max}}
    \end{equation}

\end{itemize}

\subsubsection{DP-based Task Offloading}
To address MCKP problem, we propose a DP-based approach to optimally assign image blocks to detection models across multiple edge servers. This method aims to maximize detection accuracy while ensuring that inference latency remains within the threshold $D_{\text{max}}$.

Specifically, we first determine the detection precision for each image block and the inference latency for each model. A DP table is then used to store the best accuracy values for each possible state, taking into account the maximum allowable latency $D_{\text{max}}$. During the optimization process, we evaluates all feasible model configurations and updates the optimal precision at each decision point. The final solution is derived through a backtracking procedure, which generates the optimal model offloading plan.

Based on these offloading plans, image blocks are assigned to the respective edge servers accordingly.

\section{Performance Evaluation}
\label{sec:performance evaluation}

\subsection{Experimental Setup}
\label{subsec:environment_and_parameters}

\begin{figure}[H]
    \centering
    \includegraphics[width=0.5\linewidth]{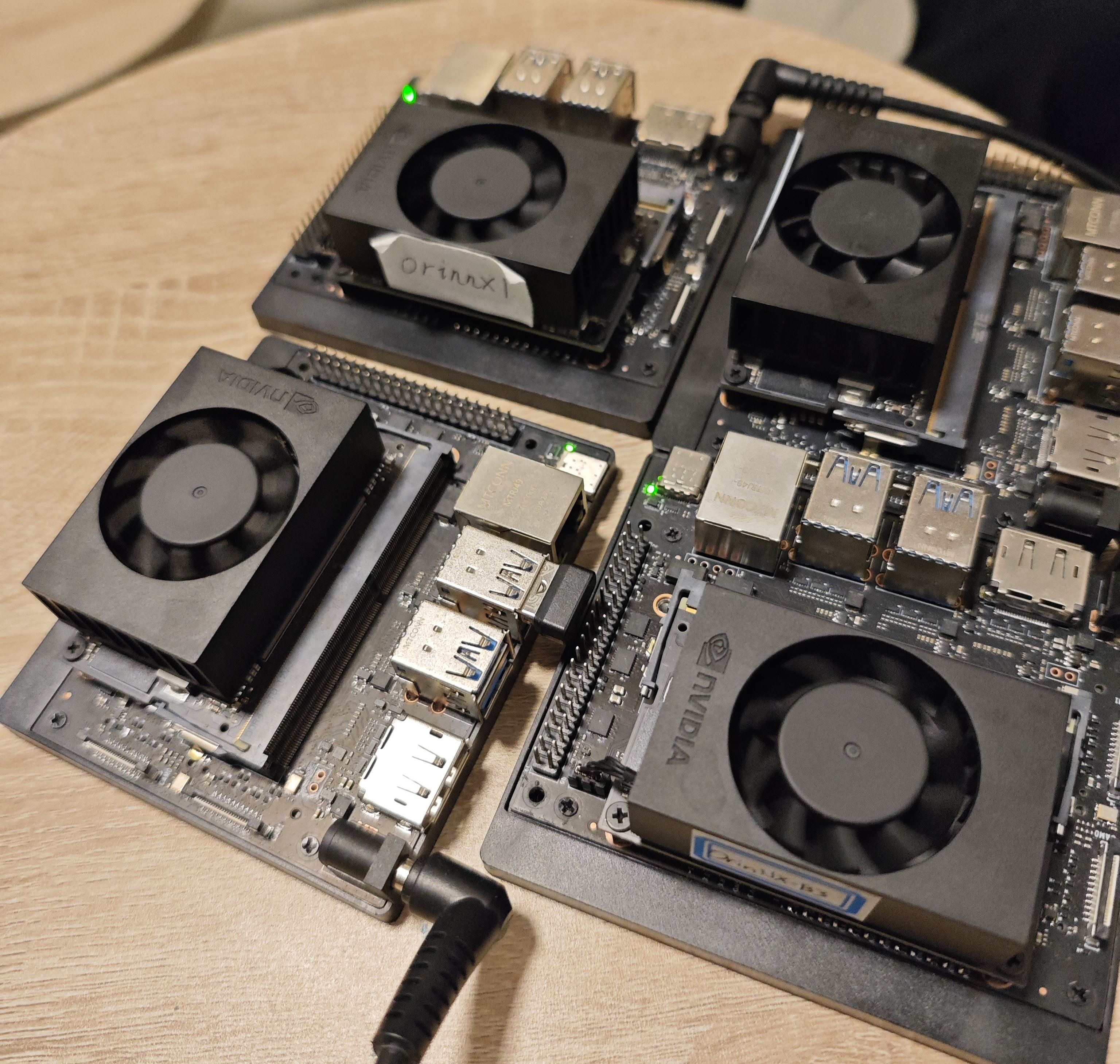}
    \caption{Experimental setup with 4 Jetson Orin}
    \label{fig:platform}
\end{figure}

\textbf{Platforms:} We implement RE-POSE on four (\(E = 4\)) NVIDIA Jetson Orin NX (16GB) devices (shown in Fig. \ref{fig:platform}). The Jetson power mode is set to MAXN to ensure maximum performance. PyTorch is utilized as the inference engine on GPUs for object detection tasks.

\begin{figure*}[th]
    \centering
    \begin{minipage}{0.3\textwidth}
        \centering
        \includegraphics[width=\textwidth]{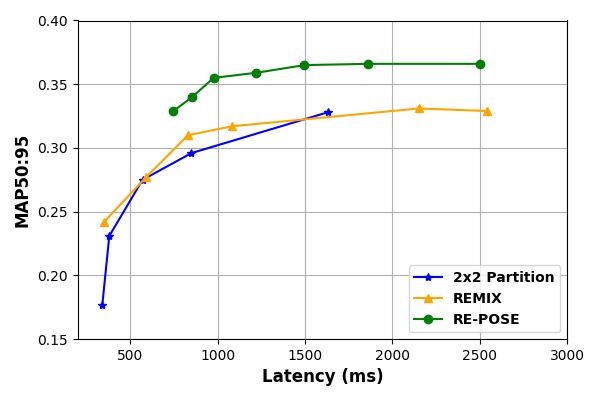}
        \caption{Inference latency and mAP50:95 on PANDA dataset}
        \label{fig:overall performance}
    \end{minipage}
    \hfill
    \begin{minipage}{0.3\textwidth}
        \centering
        \includegraphics[width=\textwidth]{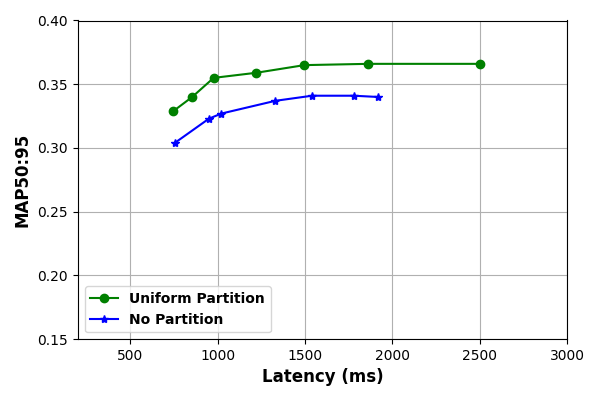}
        \caption{Performance using different ways of coarse detection}
        \label{fig:1x1 vs 2x2}
    \end{minipage}
    \hfill
    \begin{minipage}{0.3\textwidth}
        \centering
        \includegraphics[width=\textwidth]{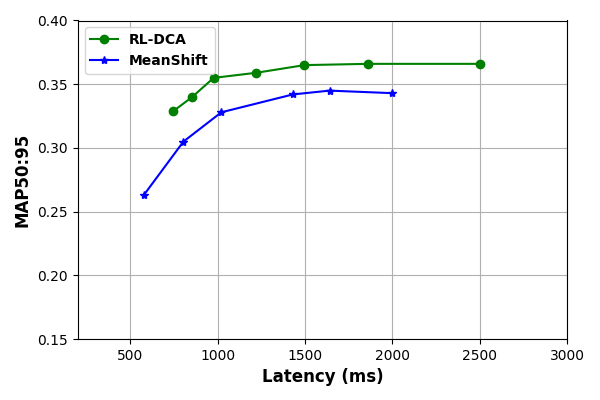}
        \caption{Performance using simple MeanShift and RL-DCA in RE-POSE}
        \label{fig:RL-DCA vs MeanShift}
    \end{minipage}
    \label{fig:main}
\end{figure*}
 
\textbf{Models:} The YOLOv8 family forms the network unit of our RE-POSE framework. In particular, we deploy different YOLOv8 models on each Jetson Orin to account for computational constraints. Specifically, each edge server is equipped with a distinct YOLOv8 model paired with the YOLOv8l model for coarse detection.

\textbf{RL-DCA parameters:} The parameters we use for RL-DCA training and inference are shown in the supplemental material. For policy network $\pi_\theta$ and critic network $V_\phi$, we use two multilayer perceptrons (MLP) consist of two hidden layers with 128 neurons each and ReLU activation functions.

\textbf{Dataset:} We evaluate RE-POSE using the PANDA \cite{wang2020panda} dataset, resizing the images from their original resolutions to 4K (\(3840 \times 2160\)).

\textbf{Baselines:} Our baselines consist of two high-resolution object detection solutions:
\begin{itemize}
    \item \textbf{Simple Partitioning with Basic Object Detection:} We uniformly partition the original image into \(2 \times 2\) blocks and perform object detection on each block independently. The results are then concatenated. An overlap ratio of 30\% is applied to prevent objects from being truncated during partitioning.
    \item \textbf{REMIX \cite{jiang2021flexible}:} REMIX is a well-established solution that enhances inference accuracy within a latency budget. It employs a prune-and-search-based method to generate partition plans, assigning an appropriate model to each block based on object distribution in historical frames.
\end{itemize}

\textbf{Metrics:} The experiments utilize mean Average Precision (mAP) at Intersection over Union (IoU) thresholds ranging from 50\% to 95\% (mAP50:95) and processing latency (\(T_l\)) as the primary evaluation metrics.

\subsection{Experimental Results}
\label{subsec:experimental results}

\subsubsection{Overall Performance} As illustrated in Fig.~\ref{fig:overall performance}, RE-POSE outperforms all baselines, achieving up to a 25\% increase in mAP50:95 under similar latency conditions. This superior performance is attributed to two main factors:

First, RE-POSE efficiently utilizes computational resources through parallel coarse detection and task offloading, enabling focused detection on critical scene areas. In contrast, Remix relies on historical frame statistics, which may not accurately capture the current object distribution.

Second, traditional partitioning methods often fragment target objects in dense scenes, leading to significant precision loss despite attempts to mitigate this with overlapping blocks. RE-POSE's RL-DCA addresses partition loss by employing one-dimensional clustering, maintaining high detection accuracy even in densely packed environments.

\subsubsection{Evaluation of Coarse Detection}
The quality of coarse detection significantly impacts overall performance. Fig.~\ref{fig:1x1 vs 2x2} shows that performing coarse detection on uniformly partitioned images (2$\times$2) consistently outperforms no partitioning under the same latency constraints. This improvement is due to better object coverage, as coarse detection on the full image misses certain objects, limiting the effectiveness of subsequent detection stages.

\subsubsection{Evaluation of RL-DCA}
Replacing RL-DCA with the MeanShift algorithm, as showed in Fig.~\ref{fig:RL-DCA vs MeanShift}, results in a noticeable decline in performance. MeanShift does not account for the correlation between object positions and sizes, leading to imbalanced clusters and inaccurate detections, especially in dense scenes. Additionally, MeanShift's inability to control the number of clusters exacerbates these issues. In contrast, RL-DCA maintains high accuracy with acceptable latency by effectively managing cluster sizes and distributions.

\section{Conclusion}
\label{sec:conclusion}
In this paper, we introduced RE-POSE, a RL-driven framework that adaptively partitions high-resolution video frames and efficiently offloads tasks among edge servers, significantly improving object detection accuracy under given latency constraints. Experimental results demonstrate that RE-POSE outperforms existing solutions for high-resolution real-time object detection.

\bibliographystyle{IEEEbib}
\bibliography{references}
\end{document}


\section{Motivation and Observations}
\label{sec:motivation and observations}
Recent advances in DNNs have enabled remarkable progress in object detection tasks. However, performing accurate low-latency inference on high-resolution video frames at the edge remains a pressing challenge. Limited computational resource, combined with the complexity of detection models and the diversity of object characteristics (like size, distribution, and presence of the object), make it difficult to simply "pick one model" for the entire frame without incurring performance penalties. In this section, we present empirical observations and insights that underscore the necessity of adapting model choices to objects across multiple size ranges.

\subsection{Correlation Between Object Size and Detection Accuracy}

Based on our preliminary experiment, we obtained the mAP of YOLOv8 for objects of different sizes, as shown in Fig. 1. According to it, we can conclude:
   \subsubsection{\textbf{Universal Improvement with Increasing Object Size}} As object size increases, the detection accuracy (mAP) systematically improves for all models. When objects are extremely small (on the order of 8–16 pixel²), none of the models—regardless of their size—achieves desirable detection performance. The mAP values approach zero in these cases, illustrating the intrinsic difficulty of small object detection.

   \subsubsection{\textbf{Enhanced Performance of Larger Models on Medium to Large Objects}} As the object size surpasses intermediate ranges (e.g., around 48–64 pixel²), accuracy values for all models begin to rise significantly. More robust models, such as YOLOv8l and YOLOv8x, consistently surpass lighter models like YOLOv8n and YOLOv8s. For instance, in the 128–160 pixel² range, YOLOv8x attains an mAP close to 0.6, whereas YOLOv8n remains around 0.35. This gap highlights that model complexity and representational capacity play a critical role when detecting moderately-sized to large objects.
   
   \subsubsection{\textbf{Plateauing for the Largest Objects}} As object sizes increase further, the improvement in mAP tends to plateau. At larger object scales, the stronger models (YOLOv8l, YOLOv8x) can achieve mAP values of approximately 0.65 or higher, whereas the lighter models cap out around 0.4–0.5. This pattern confirms that more powerful model offer a consistent advantage across the object size spectrum, most notably for large targets.

   In summary, Fig \ref{fig:map vs model} underscores that detection accuracy is inherently size-dependent and that scaling up the model generally leads to better results, especially for medium and large objects. However, even large models struggle with very small objects.

\subsection{Latency-Accuracy Trade-offs with and without Partitioning}

   \subsubsection{\textbf{Without Partitioning}}  
     Also, in our preliminary experiment, we obtain mAP and GPU inference latency of YOLOv8, as shown in Fig \ref{fig:latency vs mAP.png}. Based on it, we can can conclude:
     
     Trade-off Between Size and Performance: YOLOv8n shows the lowest latency (average latency:87.8ms) but also the lowest accuracy (mAP ~0.109). As we progress from YOLOv8n to YOLOv8x, latency increases substantially—up to about 400ms for YOLOv8x—along with a commensurate improvement in mAP (up to ~0.312). This pattern reveals a fundamental trade-off: larger models provide better detection performance but at the expense of higher computational costs and longer inference times.

   \subsubsection{\textbf{With 2x2 Partitioning}} 
     Under a 2x2 partitioning scheme, the input frame is divided into four subregions, and the detection model is applied separately to each sub-partition.
     Significantly Increased Latency for Moderate Gains in mAP: Although partitioning can improve overall detection accuracy by allowing finer granularity and potentially better detection of smaller objects, it imposes a substantial computational burden. For example, YOLOv8n’s latency under 2x2 partitioning may increase from ~87.8ms to around 340ms, and YOLOv8x’s latency can exceed 1630ms. While mAP also increases (for YOLOv8x, up to ~0.335), the soaring latency becomes a critical bottleneck for real-time applications.

   These latency-accuracy trade-offs highlight that uniformly applying a single, powerful model—either to the entire frame or uniformly partitioned subregions—yields modest accuracy benefits at a severe cost in processing time. This is particularly problematic for latency-sensitive tasks (e.g., autonomous driving, real-time surveillance), where swift decision-making is paramount.

Based on the above observations, we conclude: 1) Size-Dependent Accuracy: Larger objects are detected more accurately, and larger, more complex models (YOLOv8l, YOLOv8x) outperform their smaller counterparts across all object sizes, but particularly for medium and large objects. 2) Latent Performance Bottleneck: Although scaling up models or applying partitioning strategies can boost accuracy, they significantly increase latency, making straightforward “one-size-fits-all” solutions infeasible for real-time deployments.

These findings suggest that to achieve an optimal balance between detection accuracy and real-time responsiveness, it is insufficient to rely on uniform model selection or naive partitioning. Instead, a more nuanced, adaptive strategy is required—one that tailors model selection and inference approaches to the heterogeneous spatial and object-size distributions present in high-resolution frames.

\begin{figure}[H]
    \centering
    \includegraphics[width=0.7\linewidth]{mAP vs model.jpg}
    \caption{mAP of YOLOv8 on objects of different sizes}
    \label{fig:map vs model}
\end{figure}

\begin{figure}[H]
    \centering
    \includegraphics[width=0.9\linewidth]{latency vs mAP.png}
    \caption{mAP and GPU inference latency of YOLOv8}
    \label{fig:latency vs mAP.png}
\end{figure}

\section{Training Process of RL-DCA}
\label{pg: Training Process}

\begin{algorithm}
\caption{RL-DCA Training}
\label{alg:training}
\LinesNumbered
\KwIn{$\pi_{\theta}$: policy network, $V_{\phi}$: critic network, $E$: RL-DCA environment, $T_{\max}$: max steps per iteration, $B$: batch size, $\Gamma$: discount factor, $\epsilon$: PPO clipping parameter, $\eta_{\theta}, \eta_{\phi}$: learning rates}
\KwOut{$\pi_{\theta}$, $V_{\phi}$: trained policy and critic networks}

Initialize $\theta, \phi$ randomly\;
Initialize a trajectory memory $\mathcal{D}$\;

\For{each training iteration}{
   \textbf{Exploration Phase:}\\
   \For{$t \gets 0$ \KwTo $T_{\max}$}{
      Observe clustering state $s_t$ from $E$\;
      Sample action $a_t \sim \pi_{\theta}(a|s_t)$\;
      Execute $a_t$, observe $R_t$ and $s_{t+1}$\;
      Store $(s_t, a_t, R_t, s_{t+1})$ in $\mathcal{D}$\;
   }

   \textbf{Optimization Phase:}\\
   Compute $G_t = \sum_{k=0}^{\infty}\Gamma^k R_{t+k}$\;
   Compute $A_t = G_t - V_{\phi}(s_t)$\;

   \tcp{Update policy and value function in batches of size $B$}
   Partition $\mathcal{D}$ into mini-batches of size $B$\;
   \For{each mini-batch}{
      Compute:
      
      \ $r_t(\theta) = \frac{\pi_{\theta}(a_t|s_t)}{\pi_{\theta_{\text{old}}}(a_t|s_t)}$\;

      \resizebox{.8\linewidth}{!}{
      $L_{\text{CLIP}}(\theta) = \mathbb{E}_t\left[\min\left(r_t(\theta)A_t,\;\text{clip}(r_t(\theta),1-\epsilon,1+\epsilon)A_t\right)\right]$
      }

      \resizebox{.5\linewidth}{!}{
      $L_{\text{value}}(\phi) = \frac{1}{B}\sum_t (V_{\phi}(s_t)-G_t)^2$
      }

      \tcp{Gradient-based updates}
      Update $\theta, \phi$:
      \[
      \theta \gets \theta + \eta_{\theta} \nabla_{\theta} L_{\text{CLIP}}(\theta)
      \]
      \[
      \phi \gets \phi - \eta_{\phi} \nabla_{\phi} L_{\text{value}}(\phi)
      \]
   }
   Clear $\mathcal{D}$\;
}
\end{algorithm}

The RL-DCA training process is designed to iteratively refine the clustering policy by interacting with the environment $E$. The environment $E$ provides the agent with the current clustering configuration and transitions the state based on the agent's actions applied to the clustering configuration. The primary role of $E$ is to simulate the dynamic clustering scenario, offering agent rewards that reflect the quality of the current configuration. 

The training process (as shown in Algorithm \ref{alg:training}) includes two phases: exploration and optimization.

\textbf{Exploration}: At time step $t$, the agent observes the current clustering state $s_t$, and selects an action $a_t$ sampling from the policy $\pi_{\theta}(a_t|s_t)$. This results in a transition of the environment to state $s_{t+1}$ and the corresponding reward $R_t$. The agent stores the state-action-reward trajectory for use in policy optimization $(s_t, a_t, R_t, s_{t+1})$. This episode ends when the maximum number of steps $t_{max}$ is reached.

\textbf{Optimization}: At the end of each batch of episodes, the PPO updates the agent's policy based on the collected trajectories. PPO leverages the advantage function to quantify the beneficial effect of an action $a_t$ compared to the expected value of the current state $s_t$. The advantage is computed as:
\begin{equation}
    A_{t}=\sum_{k=0}^{\infty} \Gamma^{k} R_{t+k}-V_{\phi}\left(s_{t}\right)
\end{equation}
where the $\Gamma$ is the discount factor that prioritizes immediate rewards over future ones $(0<\Gamma<1)$, $R_{t+k}$ is the reward obtained $k$ steps after time step t, $V_\phi(s_t)$ is the value of state $s_t$ as predicted by the critic network. 

The purpose of the update of the policy network is to maximize the clipped surrogate objective $L_{clip}(\theta)$, which increases the probability of actions that lead to a greater advantage. The objective is defined as:
\begin{equation}
    L_{\text{CLIP}}(\theta) = \mathbb{E}_t \left[ \min \left( r_t(\theta) A_t, \text{clip}(r_t(\theta), 1 - \epsilon, 1 + \epsilon) A_t \right) \right]
\end{equation}
where $r_t(\theta) = \frac{\pi_\theta(a_t|s_t)}{\pi_{\theta old}(a_t|s_t)}$ is the probability ratio between the new policy $\pi_\theta$ and the old policy $\pi_{\theta old}$. clit$(r_t(\theta), 1 - \epsilon, 1+\epsilon)$ limits the ratio to the range $[1-\epsilon, 1+\epsilon]$, where $\epsilon > 0$ is a hyperparameter, which ensures that policy updates are stable and do not diverge too much from the previous policy. 

The critic network is trained simultaneously with the actor network to accurately predict the expected cumulative reward for each state. The value function update minimizes the mean squared error between the predicted value $V_\phi(s_t)$ and the actual return $G_t$:
\begin{equation}
    L_{\text{value}}(\phi) = \frac{1}{B}\sum_t(V_\phi(s_t)-G_t)^2
\end{equation}
where $G_t = \sum_{k=0}^\infty\Gamma^kR_{t+k}$ is the actual discounted cumulative return, $B$ is the number of samples in a batch.

During the optimization phase, we use gradient ascent to optimize both the policy network and the critic network. Through an iterative cycle of exploration and optimization, the policy network gradually learns to adjust the clustering configurations to align with the desired patterns.
\clearpage
\section{Task Offloading Algorithm}
\begin{algorithm}[th]
\caption{Parallel Edge Offloading}
\label{alg:parallel_edge_offloading}
\LinesNumbered
\KwIn{
    $clusters$: List of image clusters,
    $models$: Set of detection models with parameters (accuracy, inference latency),
    $D_{max}$: Maximum allowable latency (ms)
}
\KwOut{
    $Opt_{val}$: Optimal detection accuracy,
    $Plans$: Mapping of clusters to models
}

\BlankLine
\textbf{Step 1: Initialization}\\
$N \gets \text{len}(clusters)$\;
$\text{cluster\_precisions} \gets [compute\_precision(c, models) \text{ for } c \in clusters]$\;
$\text{model\_latency} \gets \{m: models[m]['latency'] \text{ for } m \in models\}$\;

$\text{dp} \gets \text{initialize\_table}(N+1, D_{max}+1, -1.0)$\;
$\text{choice} \gets \text{initialize\_table}(N+1, D_{max}+1, \text{None})$\;
\For{$t \gets 0$ \KwTo $D_{max}$}{
    $\text{dp}[0][t] \gets 0.0$\;
}

\BlankLine
\textbf{Step 2: Dynamic Programming}\\
\For{$i \gets 1$ \KwTo $N$}{
    \For{$t \gets 0$ \KwTo $D_{max}$}{
        \ForEach{$m$ in $models$}{
            \If{$t \ge \text{model\_latency}[m]$ \textbf{and} $\text{dp}[i-1][t - \text{model\_latency}[m]] + \text{cluster\_precisions}[i-1][m] > \text{dp}[i][t]$}{
                $\text{dp}[i][t] \gets \text{dp}[i-1][t - \text{model\_latency}[m]] + \text{cluster\_precisions}[i-1][m]$\;
                $\text{choice}[i][t] \gets m$\;
            }
        }
    }
}

\BlankLine
\textbf{Step 3: Retrieve optimal precision}\\
$Opt_{val} \gets \max(\text{dp}[N])$\;
$opt\_t \gets \text{argmax}(\text{dp}[N])$\;

\BlankLine
\textbf{Step 4: Backtrack to determine offloading plan}\\
$selected\_models \gets []$\;
$remaining\_t \gets opt\_t$\;
\For{$i \gets N$ \KwTo $1$}{
    $m \gets \text{choice}[i][remaining\_t]$\;
    $selected\_models.\text{append}((clusters[i-1].id, m))$\;
    $remaining\_t \gets remaining\_t - \text{model\_latency}[m]$\;
}
$selected\_models.\text{reverse}()$\;

$Plans \gets \{cid: m \text{ for } (cid, m) \text{ in } selected\_models\}$\;

\Return{$Opt_{val}$, $Plans$}\;
\end{algorithm}

\section{RL-DCA parameters:} The parameters we use for RL-DCA training and inference are shown in Table \ref{tab:RL-DCA parameter}.

\begin{table}[H]
    \centering
    \caption{RL-DCA parameters}
    \label{tab:RL-DCA parameter}
    \begin{tabular}{cc}
    \hline
        \textbf{Parameter} & \textbf{Value}\\
        \hline
       $n$  &  1 \\
       $E$  & 4 \\
       $\alpha_T$ & 0.5\\
       $N_{min}$ & 10\\
       $N_{max}$ & 15\\
       $d_m$ & 0.03\\
       $\alpha$ & 50\\
       $\beta$ & 1\\
       $\gamma$ & 1000000\\
       $\delta$ & 5\\
       $T_{max}$ & 30\\
       \hline
    \end{tabular}
    
\end{table}
\section{Latency Analysis}
Fig. \ref{fig:Latency Analysis} illustrates the latency of each processing stage in the RE-POSE system before formal detection. The stages include Coarse Detection, RL-DCA, Task Offloading, and we consider all Network Transmission as a whole stage. These pre-processing steps introduce certain challenges to the real-time performance of RE-POSE, as nearly 500ms of latency budget is consumed before the final detection begins. Selecting a smaller network for coarse detection or adjusting the RL-DCA training strategy—such as reducing the number of expected clusters and the number of selection steps—can help lower latency to some extent, but this comes at the cost of reduced accuracy. Since the PANDA dataset focuses more on dense scenes, we employ larger models and a greater number of clusters to maximize accuracy and demonstrate the superiority of our method.

\begin{figure}[H]
    \centering
    \includegraphics[width=1\linewidth]{Latency Analysis.png}
    \caption{Latency of each processing stage before final detection.}
    \label{fig:Latency Analysis}
\end{figure}